\documentclass[akbc,twoside,11pt,lettersize]{article}
\usepackage{akbc}
\usepackage[inline]{enumitem}
\usepackage{graphicx}
\usepackage{booktabs}
\usepackage{wrapfig}

\akbcheading
\ShortHeadings{CSKG}{Ilievski, Szekely, Cheng, Zhang, \& Qasemi}
 \finalcopy 
\begin{document}

\title{Consolidating Commonsense Knowledge}

\author{\name Filip Ilievski \email ilievski@isi.edu \\
      \name Pedro Szekely \email pszekely@isi.edu \\
      \addr Information Sciences Institute, University of Southern California, Marina del Rey, CA, USA
      \AND
      \name Jingwei Cheng \email chengjingwei@mail.neu.edu.cn \\
      \name Fu Zhang \email zhangfu@mail.neu.edu.cn \\
      \addr School of Computer Science and Engineering, Northeastern University, Shenyang, China
      \AND
      \name Ehsan Qasemi \email qasemi@isi.edu \\
      \addr Information Sciences Institute, University of Southern California, Marina del Rey, CA, USA}


\maketitle

\begin{abstract}
Commonsense reasoning is an important aspect of building robust AI systems and is receiving significant attention in the natural language understanding, computer vision, and knowledge graphs communities. 
At present, a number of valuable commonsense knowledge sources exist, with different foci, strengths, and weaknesses.
In this paper, we list representative sources and their properties. Based on this survey, we propose principles and a representation model in order to consolidate them into a Common Sense Knowledge Graph (CSKG). We apply this approach to consolidate seven separate sources into a first integrated CSKG. We present statistics of CSKG, present initial investigations of its utility on four QA datasets, and list learned lessons.

\end{abstract}
\section{Introduction}
\label{sec:intro}

Capturing, representing, and leveraging commonsense knowledge has been a paramount for AI since its early days, cf. \cite{mccarthy1960programs}. In the light of the modern large (commonsense) knowledge graphs and various neural advancements, the DARPA Machine Common Sense program \cite{gunning2018machine} represents a new effort to understand commonsense knowledge through question-answering evaluation benchmarks. An example of such question from the SWAG dataset~\cite{zellers2018swag} describes a woman that takes a sit at the piano: 

\footnotesize{
\begin{verbatim}
                   Q: On stage, a woman takes a seat at the piano. She:
                      1. sits on a bench as her sister plays with the doll.
                      2. smiles with someone as the music plays.
                      3. is in the crowd, watching the dancers.
                   -> 4. nervously sets her fingers on the keys.
\end{verbatim}
}

\normalsize

Realizing that the logical next step is her ``nervously setting her fingers on the keys'' is out of reach for typical information retrieval strategies, as there is no lexical overlap between the situation and the correct answer. Although language models \cite{devlin2018bert,liu2019roberta} capture linguistic patterns that allow them to perform well on many questions, they have no mechanism to fill gaps of knowledge in communication.\footnote{Recent work provides evidence that language models, while probably useful and often impressive, are non-robust when applied to semantic tasks, lacking mechanisms to understand plausibility or keep track of evolving states of events and entities~\cite{marcus2020next}. They struggle with higher number of inference steps \cite{richardson2019does}, role-based event prediction \cite{ettinger2020bert}, as well as numeric, emotional, and spatial inference \cite{bhagavatula2019abductive}. On the other hand, systems like KagNet~\cite{lin2019kagnet} and HyKAS \cite{ma2019towards} have managed to enhance language models by combining them with background knowledge from ConceptNet~\cite{speer2017conceptnet}}.
Filling such gaps requires a more complex, situational reasoning, for which the language models need to be enriched with suitable background knowledge, as in \cite{lin2019kagnet}. 

Intuitively, graphs of (commonsense) knowledge contain such background knowledge that humans possess and apply, but machines cannot access or distill directly in communication. A number of such knowledge sources exist today, which presents a unique opportunity for reasoning in downstream tasks. Taxonomies, like WordNet \cite{miller1995wordnet}, organize conceptual knowledge into a hierarchy of classes. An independent ontology, coupled with rich instance-level knowledge, is provided by Wikidata \cite{vrandevcic2014wikidata}, a structured version of Wikipedia. FrameNet \cite{baker1998berkeley}, on the other hand, defines an orthogonal structure of frames and roles; each of which can be filled with a WordNet/Wikidata class or instance. Sources like ConceptNet \cite{speer2017conceptnet} or WebChild \cite{tandon2017webchild}, provide more `episodic' commonsense knowledge, whereas ATOMIC \cite{sap2019atomic} captures pre- and post-situations for an event. Finally, image description datasets, like Visual Genome \cite{krishna2017visual}, have visual commonsense knowledge.

Considering the above example, ConceptNet's triples state that pianos have keys and are used to perform music, which supports the correct option and discourages answer 2. WordNet states specifically, though in natural language, that pianos are played by pressing keys. According to an image description in Visual Genome, a person could play piano while sitting and having their hands on the keyboard. In natural language, ATOMIC indicates that before a person plays piano, they need to sit at it, be on stage, and reach for the keys. ATOMIC also lists strong feelings associated with playing piano. FrameNet's frame of a performance contains two separate roles for the performer and the audience, meaning that these two are distinct entities, which can be seen as evidence against answer 3.
While these sources clearly provide complementary knowledge that can help commonsense reasoning, their representation formats, principles and foci are different, making integration difficult.

In this paper, we propose an approach for integrating these (and more sources) into a single Common Sense Knowledge Graph (CSKG). We start by surveying existing sources of commonsense knowledge to understand their particularities (section \ref{sec:sources}). We summarize key challenges and related efforts on consolidating commonsense knowledge in section \ref{sec:integration}. Based on the survey and the listed challenges, we devise five principles and a representation model for a consolidated CSKG (section \ref{sec:approach}). In section \ref{sec:cskg} we apply our approach to build the first version of CSKG, by combining seven complementary, yet disjoint, sources. Here, we also compare the evidence provided by CSKG compared to ConceptNet on four commonsense QA datasets. In section \ref{sec:discussion} we reflect on CSKG and discuss its role in future research. We conclude the paper in section \ref{sec:conclusions}.



\begin{table}[!t]
	\centering
	{\footnotesize
	\caption{Survey of existing sources of commonsense knowledge.}
	\begin{tabular} {p{1.6cm} p{2.3cm} p{1.5cm} p{3cm} p{1.7cm} p{2.8cm}}
		\toprule
		  & \bf describes & \bf creation & \bf size & \bf mappings & \bf examples \\
		\midrule
				
		\bf Concept Net & everyday objects, actions, states, relations (multilingual) & crowd-sourcing & 36 relations, 8M nodes, 21M edges & WordNet, DBpedia, OpenCyc, Wiktionary & \texttt{/c/en/piano} \texttt{/c/en/piano/n} \texttt{/c/en/piano/n/wn} \texttt{/r/relatedTo}\\ \hline
		\bf Web Child & everyday objects, actions, states, relations & curated automatic extraction & 4 relation groups, 2M nodes, 18M edges & WordNet & \texttt{hasTaste} \texttt{fasterThan}  \\ \hline
		\bf ATOMIC & event pre/post-conditions & crowd-sourcing & 9 relations, 300k nodes, 877k edges & ConceptNet, \hspace{9pt} Cyc & \texttt{wanted-to} \texttt{impressed} \\ \hline
		\bf Wikidata & instances, concepts, relations & crowd-sourcing & 1.2k relations, 75M objects, 900M edges & various & \texttt{wd:Q1234} \texttt{wdt:P31} \\ \hline
		\bf CEO & event pre/ post-conditions & manual & 121 properties, 223 events & FrameNet, SUMO & \texttt{ceo:Damaging} \texttt{hasPostSituation} \\ \hline
		\bf WordNet & words, concepts, relations & manual & 10 relations, 155k words, 176k synsets & & \texttt{dog.n.01} \texttt{hypernymy}    \\ \hline
        \bf Roget & words, relations & manual & 2 relations, 72k words, 1.4M edges & & \texttt{truncate}\hspace{9pt} \texttt{antonym} \\ \hline
		\bf VerbNet & verbs,\hspace{9pt} relations & manual & 273 top classes  23 roles,\hspace{14pt} 5.3k senses & FrameNet, WordNet & \texttt{perform-v} \texttt{performance-26.7-1} \\ \hline
		\bf FrameNet & frames, roles, relations & manual & 1.9k edges, 1.2k frames, 12k roles, 13k lexical units & & \texttt{Activity} \texttt{Change\_of\_leadership} \texttt{New\_leader} \\ \hline
		\bf Visual Genome & image objects, relations, attributes & crowd-sourcing & 42k relations, 3.8M nodes, 2.3M edges, 2.8M attributes & WordNet & \texttt{fire hydrant} \texttt{white dog} \\\hline
		\bf ImageNet & image objects & crowd-sourcing & 14M images, 22k synsets & WordNet & \texttt{dog.n.01} \\ \hline
		\bf Flickr 30k & image\hspace{9pt} objects & crowd-sourcing & 30k images, 750 objects & & \texttt{her backyard} \hspace{9pt}\texttt{red bags} \\
		\bottomrule
	\end{tabular}
}
	\label{tab:survey}
\end{table}

\section{Sources of Common Sense Knowledge}
\label{sec:sources}

We survey existing commonsense knowledge sources: \textbf{ConceptNet}~\cite{speer2017conceptnet}, \textbf{WebChild}~\cite{tandon2017webchild}, \textbf{ATOMIC}~\cite{sap2019atomic}, \textbf{Wikidata}~\cite{vrandevcic2014wikidata}, \textbf{CEO}~\cite{segers2018circumstantial}, \textbf{WordNet}~\cite{miller1995wordnet}, \textbf{Roget}~\cite{kipfer2005roget}, \textbf{VerbNet}~\cite{schuler2005verbnet}, \textbf{FrameNet}~\cite{baker1998berkeley}, 
\textbf{Visual Genome}~\cite{krishna2017visual}, \textbf{ImageNet}~\cite{deng2009imagenet}, and \textbf{Flickr30k}~\cite{plummer2016flickr30k}.\footnote{Labels that refer to the same image object in Flickr30k were clustered by \citet{miltenburg2016stereotyping}.} 
Table \ref{tab:survey} summarizes their content, creation method, size, external mappings, and example resources. 

Primarily, we observe that the commonsense knowledge is spread over a number of sources with different focus: commonsense knowledge graphs (e.g., ConceptNet), general-domain knowledge graphs (e.g., Wikidata), lexical resources (e.g., WordNet, FrameNet), taxonomies (e.g., Wikidata, WordNet), and visual datasets (e.g., Visual Genome). Therefore, these sources together cover a rich spectrum of knowledge, ranging from everyday knowledge, through event-centric knowledge and taxonomies, to visual knowledge. While the taxonomies have been created manually by experts, most of the commonsense and visual sources have been created by crowdsourcing or curated automatic extraction.\footnote{Commonsense subsets of existing knowledge sources are sometimes also included, e.g., ConceptNet reuses knowledge from Wiktionary and DBpedia.} Similarly, commonsense and general knowledge graphs tend to be relatively large, with millions of nodes and edges; whereas the taxonomies and the lexical sources are notably smaller. Despite the diverse nature of these sources 
, we observe that many contain mappings to WordNet, as well as a number of other sources. These mappings might not be complete, e.g., only a small portion of ATOMIC can be mapped to ConceptNet and less than 1\% to Cyc. Nevertheless, these high-quality mappings provide an opening for consolidation of commonsense knowledge, a goal we pursue in this paper.

\section{Problem Statement}
\label{sec:integration}

Combining such commonsense knowledge sources in a single graph faces three key challenges.

Firstly, the sources follow \textbf{different knowledge modeling approaches}. One such difference concerns the relation set: there are very few relations in ConceptNet and WordNet, but (tens of) thousands of them in Wikidata and Visual Genome. Consolidating these sources then inherently requires a decision on how to model the relations. The granularity of knowledge is another factor of variance. While regular RDF triples fit some sources (e.g., ConceptNet), representing entire frames (e.g., in FrameNet), event conditions (e.g., in ATOMIC), or compositional image data (e.g., Visual Genome) might benefit from a more open format. An ideal representation would support the entire spectrum of granularity.

Secondly, as a number of these sources have been created to support natural language applications, they often contain \textbf{imprecise descriptions}. Natural language phrases are often the main node types in the provided knowledge sources, which provides the benefit of easier access for natural language algorithms, but it introduces ambiguity which might be undesired from a formal semantics perspective. An ideal representation would consolidate various phrasings that share a concept or a referent, while still allowing easy and efficient access to these concepts based on their natural language labels or aliases. 

Thirdly, although these sources contain links to existing ones, we observe \textbf{sparse overlap}. As these external links are typically to WordNet, and vary in terms of their version (3.0 or 3.1) or target (lemma or synset), 
the sources are still disjoint and establishing (identity) connections is difficult. Bridging these gaps, through optimally leveraging existing links, or extending them with additional ones automatically, is a modeling and integration challenge.

Previous efforts that combine commonsense resources exist. A unidirectional manual mapping from VerbNet classes to WordNet and FrameNet is provided by the Unified Verb Index~\cite{trumbo2006increasing}. The Predicate Matrix~\cite{de2016predicate} has a full automatic mapping between lexical resources, including FrameNet, WordNet, and VerbNet. PreMOn~\cite{corcoglioniti2016premon} formalizes these in RDF. \citet{miller2014wordnet,mccrae2018mapping} produce partial mappings between WordNet and Wikipedia/DBpedia. Recent systems integrate parts of these sources in an ad-hoc manner to reason on a downstream task, e.g., \citet{zareian2020bridging} combine edges from Visual Genome, WordNet, and ConceptNet in a neural network that produces a scene graph from an image.

\section{Approach}
\label{sec:approach}

To address the aforementioned challenges, we devise principles and a respective representation format that are driven by: \textbf{simplicity}, \textbf{modularity}, and \textbf{utility}. It should be simple to integrate the graph represented in this format and its arbitrary subsets in reasoning systems, compute (graph and word) embeddings, and run off-the-shelf link prediction tools.



\subsection{Principles}

We propose that the construction of a unified CSKG should follow five principles:

\noindent \textbf{P1. Embrace heterogeneity of nodes} While building CSKG, one should preserve the natural node diversity inherent to the variety of sources considered. This entails 
    blurring the distinction between objects (such as those in Visual Genome or Wikidata), classes (such as those in WordNet or ConceptNet), words (in Roget), actions (in ATOMIC or ConceptNet), frames (in FrameNet), and states (as in ATOMIC). It also allows formal nodes, describing unique objects, to co-exist with fuzzy nodes describing ambiguous lexical expressions.

\noindent \textbf{P2. Reuse edge types across resources} To support reasoning algorithms, edge types should be kept to minimum and reused across resources wherever possible. 
    For instance, the ConceptNet edge type \texttt{/r/RelatedTo} could be reused to relate a Visual Genome object (e.g., `piano') to its attributes (e.g., `black' or `room'). Note that we do not propose to impoverish the semantics of existing relations.
    
\noindent \textbf{P3. Leverage external links} The separate graphs are mostly disjoint according to their formal knowledge. However, high-quality links may exist or may be easily inferred, in order to connect these graphs and enable path finding. For instance, while ConceptNet and Visual Genome do not have direct connections, they both make reference to WordNet synsets. Investing an effort in aligning these WordNet synsets would produce a number of very valuable connections between the two knowledge sources.

\noindent \textbf{P4. Generate high-quality probabilistic links} Experimenting with inclusion of additional probabilistic links would be beneficial, as it would combat sparsity and help path finding algorithms reason over CSKG. These could be inferred with off-the-shelf link prediction algorithms, or with specialized algorithms (see section \ref{sec:cskg} for an example).

\noindent \textbf{P5. Enable access to labels} The text typically associated with KG entities, like labels or aliases, provides application-friendly and human-readable access to the CSKG. It can also help us unify descriptions of the same/similar concept across sources. We need to ensure that the graph format supports easy and efficient natural language queries over this text.  

\subsection{Representation}

We model CSKG as a \textbf{property graph}, consisting of nodes and edges information in a tabular format. 
We opted for this representation rather than the traditional RDF/OWL2 because it allows us to fulfill our goals (of simplicity and utility) and follow our principles more directly, without compromising on the format. For instance, natural language access (principle P5) to RDF/OWL2 nodes requires graph traversal over its \texttt{rdfs:label} relations. Including both reliable and probabilistic nodes would require a mechanism to easily indicate edge weights, which in RDF/OWL2 entails inclusion of blank nodes, and a number of additional edges. Moreover, the simplicity of our tabular format allows us to use standard off-the-shelf functionalities and mature tooling, like the \texttt{pandas}\footnote{\url{https://pandas.pydata.org/}} and \texttt{graph-tool}\footnote{\url{https://graph-tool.skewed.de/}} 
libraries in Python. We can also compute embeddings with \cite{lerer2019pytorch} or plug CSKG in a reasoner, e.g., \cite{wang2018yuanfudao}, with minimal adaptation, as these expect tabular inputs.

\noindent \textbf{Node representation} Each node is described by six columns: \texttt{id} (reused from the source data when possible), \texttt{label} (its primary label), \texttt{aliases} (additional labels), \texttt{pos} (part-of-speech tag, if applicable), \texttt{datasource} (original source, one of: `cn', `vg', `wn`, `rg', `wd`, `fn', `at', or `mowgli' for custom nodes), and \texttt{other} (a dictionary with provenance information, e.g., `image\_id'). Due to its generality, this node representation is suitable for any kind of node: object, state, class, or action, thus satisfying our principle P1. To accommodate P5, we ensure easy access to node labels and aliases by assigning them dedicated columns.

\noindent \textbf{Edge representation} Edges are described by six columns: \texttt{subject} (the subject ID of the edge), \texttt{predicate} (edge label), \texttt{object} (its object ID), \texttt{datasource} (its original source: `cn', `vg', `wn', `rg', `wd', `fn', `at', or `mowgli' for custom relations), \texttt{weight} (as given by the source), and \texttt{other} (provenance information, e.g., source sentence). 
Following the approach of ConceptNet and our principle P2, we minimize the set of edges. We introduce a dedicated relation, \texttt{mw:SameAs}, to indicate identity. When identity between two nodes is readily available in an external source, e.g., as a mapping, we create a link with a weight of 1.0 (principle P3). Per P4, we also predict implicit edges with weights between 0 and 1.

\begin{figure}
    \includegraphics[width=0.68\textwidth]{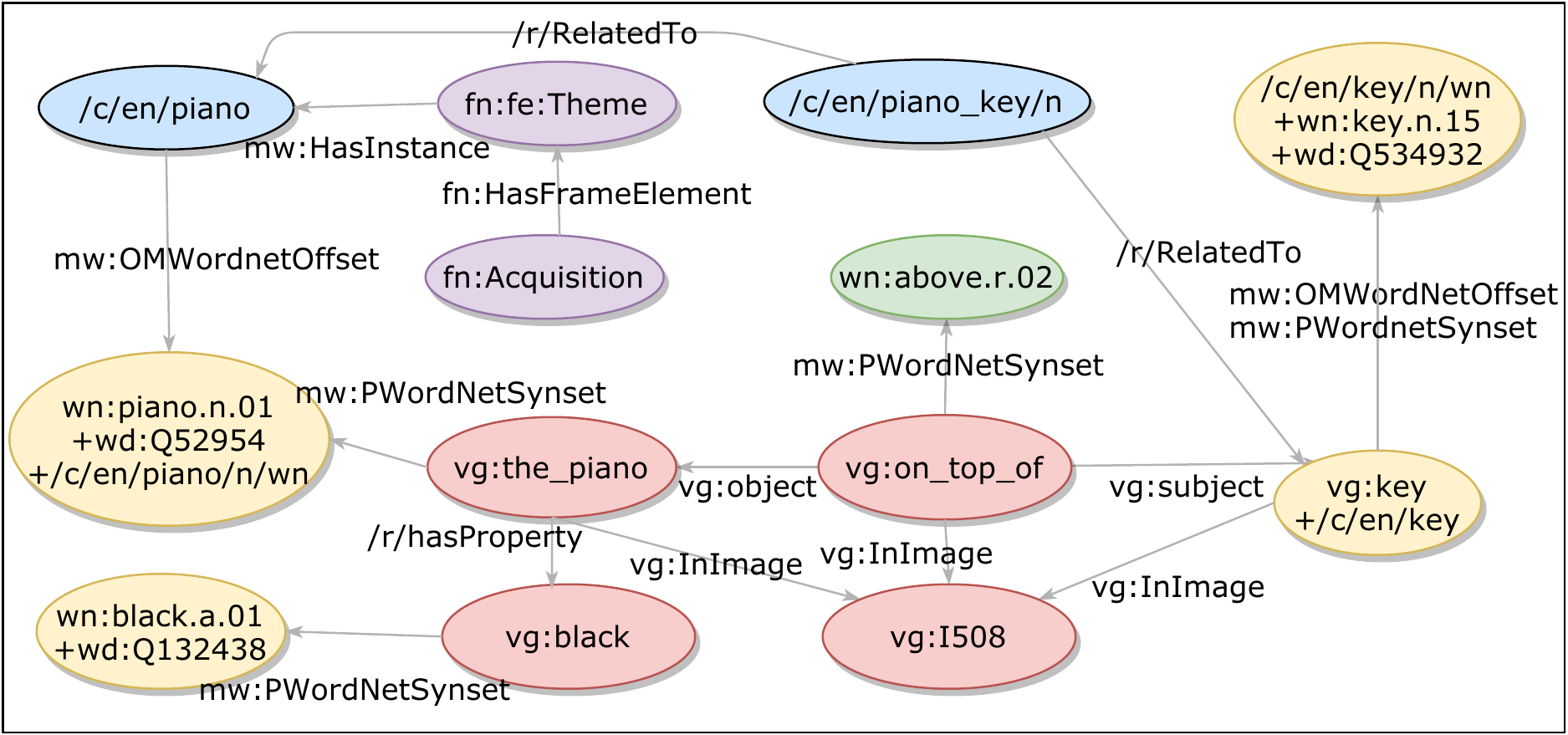}
    \centering
    \caption{Snippet from the Common Sense Knowledge Graph (CSKG). The yellow nodes are created by merging nodes from individual resources, e.g., \texttt{vg:key+/c/en/key} combines one lexical expression from Visual Genome and one from ConceptNet.}
    \label{fig:cskg}
    \vspace{-0.2in}

\end{figure}

\section{The Common Sense Knowledge Graph}
\label{sec:cskg}

In its current form (Figure \ref{fig:cskg}), CSKG integrates seven sources: a commonsense knowledge graph ConceptNet, a visual commonsense source Visual Genome, a procedural source ATOMIC, a general-domain source Wikidata, and three lexical sources, WordNet, Roget, and FrameNet. 
Here, we present our design decisions per source, their integration, and statistics on the resulting CSKG graph. We also present initial investigations on utilizing CSKG for commonsense question answering.

\subsection{Consolidation}

\subsubsection{Individual sources}

\noindent \textbf{ConceptNet} We keep its original data and representation, and we increase its consistency and connectivity in two ways. Firstly, we compute closure over the seven symmetric relations defined in ConceptNet as the symmetry was not consistently reflected in the data. Secondly, we observe that different forms of a concept are not formally linked. 
We introduce two mutually inverse relations, \texttt{mw:POSForm} and \texttt{mw:IsPOSFormOf}, to link a lemma to its part-of-speech version. Also, we define a \texttt{mw:PartOfSpeech} class, and we add edges of type \texttt{mw:OMWordnetOffset} between a lemma node and its WordNet v3.1 offset form. 

\noindent \textbf{Visual Genome} does not have its data formatted in a Semantic Web compliant format. We follow the ConceptNet approach and represent the label of each object, relation, and attribute as a node in the graph (e.g., \texttt{vg:dog}). We introduce two relations \texttt{vg:Subject} and \texttt{vg:Object} to indicate the subject and the object of a relation node. We also establish direct, symmetric links between a subject and an object node by reusing the relation \texttt{/r/RelatedTo} from ConceptNet. This relation is also leveraged to represent relations between nodes and their properties. To contextualize an object/relation in terms of its image, we create nodes for image objects (prefix \texttt{vg:I*}) and utilize the relation \texttt{vg:InImage}. For each object, relation, or attribute, we use the relation \texttt{mw:PWordnetSynset} to connect it to its WordNet v3.0 synset provided in the data dump. 


\noindent \textbf{WordNet} We include hypernymy from WordNet v3.0, via the \texttt{rdfs:subClassOf} relation.

\noindent \textbf{Roget} We include all synonyms and antonyms between words in Roget, by reusing the Concept relations \texttt{/r/Synonym} and \texttt{/r/Antonym}.

\noindent \textbf{ATOMIC} We include the entire knowledge graph, preserving the original nodes and relations. To enhance lexical matching over the labels in CSKG, we normalize the labels of the events and their attributes: converting them to lowercase, removing references to `Person*', and excluding `none' values.

\noindent \textbf{Wikidata} We include the Wikidata taxonomy through the \texttt{rdfs:subClassOf} relation.

\newcommand{\framenet}{FrameNet }
\newcommand{\cskg}{CSKG }
\newcommand{\conceptnet}{ConceptNet }
\newcommand{\wordnet}{WordNet }

\noindent \textbf{FrameNet} 
Four node types from the FrameNet ontology are imported into \cskg: Frames, Frame Elements (FEs), Lexical units (LUs), and Semantic Types (STs). We reuse 5 categories of FrameNet edges: Frame-Frame (13 edge types), Frame-FE (1 edge type), Frame-LU (1 edge type), FE-ST (1 edge type), and ST-ST (3 edge types). 

\subsubsection{Mappings}

\noindent \textbf{WordNet-WordNet} The WordNet v3.1 offsets in ConceptNet and the WordNet v3.0 synsets from Visual Genome are aligned by leveraging ILI: the WordNet InterLingual Index.\footnote{\url{https://github.com/globalwordnet/ili}} The generated 117,097 mappings are expressed through our identity relation, \texttt{mw:SameAs}.

\noindent \textbf{WordNet-Wikidata} We compute probabilistic links between WordNet synsets and Wikidata taxonomy nodes. 
Our approach consists of three components: a Candidate Retrieving Module (CRM), a Similarity Calculating Module (SCM), and a Mapping Module (MM). 
CRM retrieves candidate nodes from a customized ElasticSearch index of Wikidata. 
Concretely, it matches a synset word in any text field (including labels, aliases, and descriptions), and it ranks the candidates with a version of the default TF-IDF-based algorithm which adapts the score proportionally to the number of incoming links.\footnote{Documentation: \url{shorturl.at/jpBY3}} The top-ranked $n=50$ candidates are retained.
Then, SCM computes sentence embeddings of the descriptions of the  WordNet synset and each of the Wikidata candidates by using a pre-trained XLNet model \cite{yang2019xlnet}. The similarity between a synset and a Wikidata node is computed as a cosine similarity between their corresponding embeddings. 
MM creates a \texttt{mw:SameAs} edge between a WordNet synset and the Wikidata candidate with highest similarity. This similarity is represented as weight of the mapping edge. 
The accuracy of each mapping has been validated by one student. In total, 17 students took part in this validation. Out of all edges produced by the algorithm (112,012), the manual validation marked 57,145 as correct. We keep these in CSKG and discard the rest.

\noindent \textbf{FrameNet-ConceptNet} We connect the \framenet nodes to ConceptNet in two ways. Its lexical units are mapped to corresponding \conceptnet nodes through the Predicate Matrix (cf. section~\ref{sec:integration}), producing $3,016$ \texttt{mw:SameAs} edges.\footnote{During this step, we manually fixed approximately $250$ nodes which contained spelling and tokenization errors in the Predicate Matrix.}
Then, we use $\approx 200k$ hand-annotated sentences from the \framenet corpus, each annotated with its target frame, a set of FEs, and the words associated with each FE. We consider the set of words to be an instance of that specific FE. We ground these sets of words to \conceptnet with the rule-based method of \cite{lin2019kagnet}, thus adding 45,659 \texttt{mw:HasInstance} edges. 

\noindent \textbf{Roget-ConceptNet} We establish 60,307 \texttt{mw:SameAs} relations between word-representing nodes in Roget and in ConceptNet by a simple lexical match of their labels.

\noindent \textbf{Visual Genome-ConceptNet} We establish 32,283 \texttt{mw:SameAs} relations between lexical nodes in Visual Genome and in ConceptNet by exact matching over their labels.

\noindent \textbf{ATOMIC-ConceptNet} We establish 14,272 \texttt{mw:SameAs} relations between lexical nodes in ATOMIC and in ConceptNet by an exact match of their labels.

\subsubsection{Refinement} 

We consolidate the seven sources and six mappings as follows. Firstly, we \textbf{deduplicate} each edge table by aggregating over its first three columns (\texttt{subject, predicate, object}) and combining the values for the remaining columns. Similarly, we deduplicate each node table by aggregating over its \texttt{id} column and combining the values for the other columns. Secondly, we \textbf{concatenate} all 13 edge tables into one, same for the 6 node tables, to produce a raw version of CSKG. Thirdly, we \textbf{merge} identical nodes, by combining nodes that are connected with a \texttt{mw:SameAs} link. This operation is reflected in both the nodes and the edges table. The deduplicated version of the result is our consolidated CSKG.

\begin{table}[!t]
	\centering
	{\footnotesize
	\caption{CSKG statistics. Abbreviations: CN=ConceptNet, VG=Visual Genome, WN=WordNet, RG=Roget, WD=Wikidata, FN=FrameNet, AT=ATOMIC.}
	\begin{tabular} {l c c c c c c c c}
		\toprule
		  & \bf CN & \bf VG & \bf WN & \bf RG & \bf WD & \bf FN & \bf AT & \bf CSKG \\
		\midrule
				
		\# nodes & 1,787,276  & 316,660 & 87,942 & 71,804 & 2,388,479 & 36,582 & 288,943 & \bf 4,738,502 \\
		\# edges & 7,211,322 & 4,833,879 & 89,089 & 1,403,461 & 2,926,639 & 79,060 & 704,315 & \bf 17,210,065 \\
		max degree & 587,358 & 127,580 & 404 & 1,549 & 964,400 & 2,438 & 9,856 & \bf 964,400 \\
		mean degree & 8.07 & 30.53 & 2.03 & 39.09 & 2.45 & 4.32 & 4.87 & \bf 7.26 \\
		std degree & 0.34 & 0.83 & 0.02 & 0.34 & 0.58 & 0.16 & 0.06 & \bf 0.32 \\
		\bottomrule
	\end{tabular}
}
	\label{tab:statistics}
\end{table}

\begin{figure}
    \includegraphics[width=\textwidth]{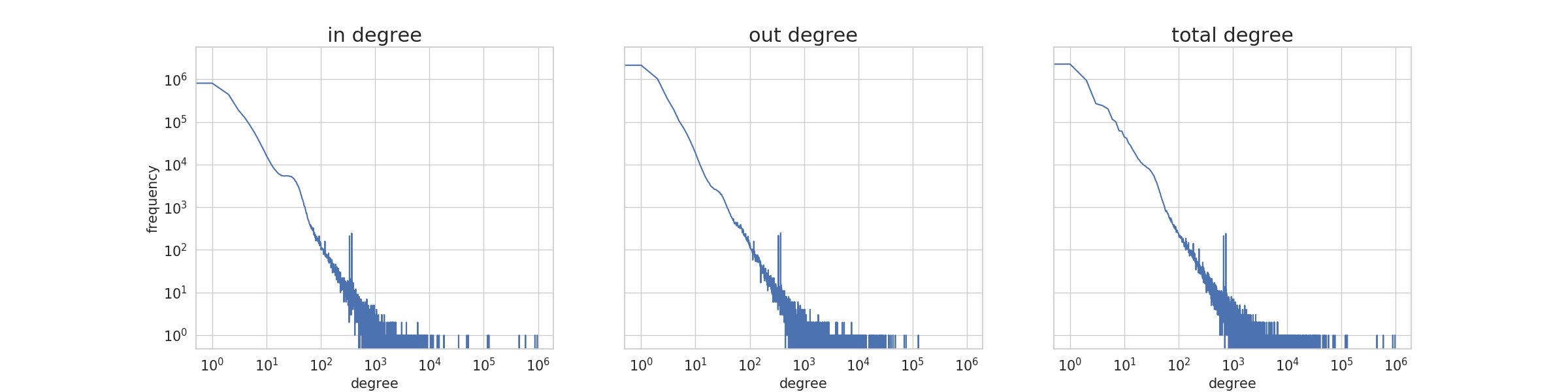}
    \centering
    \caption{Degree distribution (log-log plots).}
    \label{fig:degrees}
    \vspace{-0.15in}
\end{figure}

\subsection{Statistics}


\textbf{Basic statistics} are shown in Table 2. 
In total, our mappings produce 284,121 \texttt{mw:SameAs} links. A small portion (less than $1\%$) of the nodes and edges were duplicates. After refinement, i.e., removal of the duplicates and merging of the identical nodes, CSKG consists of 4.7 million nodes and 17.2 million edges. 
In terms of edges, its largest subgraph is ConceptNet (7.2 million), whereas Visual Genome comes second with 4.8 million edges. Wikidata contributes with the largest number of nodes, closely followed by ConceptNet. The three most common relations in CSKG are: \texttt{/r/RelatedTo} (4.6 million), \texttt{vg:InImage} (3 million), and \texttt{rdfs:subClassOf} (3 million).


\textbf{Degree distribution} The mean degree of CSKG, after merging identical nodes, grows from 7.00 to 7.26. 
Its standard deviation is 0.32, similar to ConceptNet. 
The best connected subgraphs are Roget and Visual Genome, with mean degrees of 39.09 and 30.53, respectively. The least connected graphs are the hierarchies of WordNet and Wikidata.
The degrees of WordNet and ATOMIC are fairly uniform across their nodes, whereas Visual Genome and Wikidata have higher variation. The large difference between the variations of WordNet and Wikidata can be explained by the fact that WordNet nodes typically have a single parent, whereas the Wikidata ontology is more flat and has many arcs from its leaves to high-level nodes. The maximum degree in CSKG is nearly a million, which is due to Wikidata.

\begin{wrapfigure}{r}{0.43\textwidth}
\includegraphics[width=0.4\textwidth]{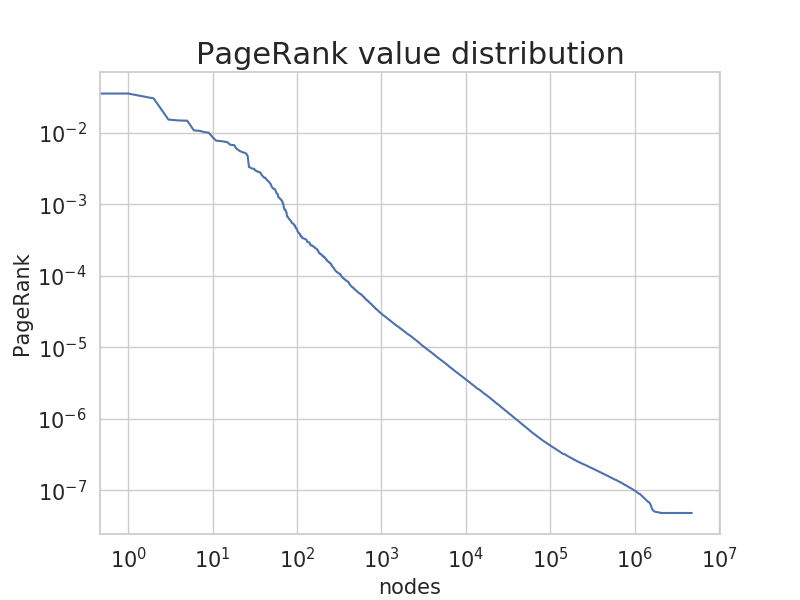}
    \centering
    \caption{PageRank distribution.}
    \label{fig:pagerank}
\end{wrapfigure}

Figure \ref{fig:degrees} presents the input, output, and total degree distributions. The highest out-degree is 10 times lower than the highest in-degree. Although most nodes have a low degree and the frequency decreases for the higher degrees, nearly $10^5$ nodes still have a total degree of 10, and many others have a much higher one. This indicates that CSKG is a well-connected graph.

\textbf{Centrality} We compute PageRank and HITS metrics of CSKG. Its PageRank distribution (Figure \ref{fig:pagerank}) indicates that while most nodes have a low PageRank value, $\approx100$ have PageRank of over 0.001, The top PageRank nodes are: \texttt{wn:polypeptide.n.01+/c/en/polypeptide/n/wn/substance+wd:Q8054} (polypeptide as a noun, merged from three sources), \texttt{wd:Q7187} (gene), and \texttt{wd:Q20747295} (protein-coding gene). According to HITS, the hubs in CSKG are: \texttt{wd:Q20747295} (protein-coding gene), \texttt{wd:Q7187} (gene), and \texttt{wd:Q427087} (non-coding RNA). The top authorities are also Wikidata nodes. The dominance of Wikidata in the centrality metrics is due to its taxonomy, where subclass relations are often directed at high-level nodes. It is unclear, however, whether such bio-informatics nodes hold value for commonsense reasoning. Future work should investigate which subsets of Wikidata contain commonsense knowledge.

\subsection{CSKG on Downstream Tasks}
\label{ssec:downstream}

\begin{table}[!t]
	\centering
	{\footnotesize
	\caption{Number of triples retrieved with ConceptNet (CN), CSKG's ConceptNet subset (CSKG-CN), and CSKG on different datasets. \#Q=number of questions.}
	\begin{tabular} {l | c c c c | c c c c}
		\toprule
		  & \multicolumn{4}{c}{\emph{train}} & \multicolumn{4}{c}{\emph{dev}} \\
		  & \bf \#Q & \bf CN & \bf CSKG-CN & \bf CSKG & \bf \#Q & \bf CN & \bf CSKG-CN & \bf CSKG \\
		\midrule
		CSQA  & 9,741 & 78,729 & 106,619 & 153,442 & 1,221 & 9,758 & 13,132 & 19,036 \\
		 \midrule
		SIQA & 33,410 & 126,596 & 189,859 & 330,200 & 1,954 & 7,850 & 11,654 & 19,953  \\
		 \midrule
		PIQA & 16,113 & 18,549 & 28,996 & 70,131 & 1,838 & 2,170 & 3,401 & 8,071 \\
		 \midrule
		aNLI & 169,654 & 257,163 & 389,640 & 771,318 & 1,532 & 5,603 & 8,477 & 16,456 \\
		
		\bottomrule
	\end{tabular}
}
    \vspace{-0.1in}
	\label{tab:downstream}
\end{table}

In a preliminary investigation, we measure the relevance of CSKG for commonsense question answering tasks, by comparing the number of retrieved triples that connect keywords in the question and in the answers. For this purpose, we adapt the lexical grounding in HyKAS~\cite{ma2019towards} to retrieve triples from CSKG instead of its default knowledge source, ConceptNet. We expect that CSKG can provide much more evidence than ConceptNet, both in terms of number of triples and their diversity. We experiment with four commonsense datasets: CommonSense QA (CSQA)~\cite{talmor2018commonsenseqa}, Social IQA (SIQA)~\cite{sap2019socialiqa}, Physical IQA (PIQA)~\cite{bisk2019piqa}, and abductive NLI (aNLI)~\cite{bhagavatula2019abductive}. As shown in Table \ref{tab:downstream}, CSKG significantly increases the number of evidence triples that connect terms in questions with terms in answers, in comparison to ConceptNet. We note that the increase on all datasets is roughly three-fold, the expected exception being CSQA, which was inferred from ConceptNet.

We inspect a sample of questions to gain insight into whether the additional triples are relevant and could benefit reasoning. For instance, let us consider the CSQA question ``Bob the lizard lives in a warm place with lots of water.  Where does he probably live?'', whose correct answer is ``tropical rainforest''. In addition to the ConceptNet triple \texttt{/c/en/lizard /c/en/AtLocation /c/en/tropical\_rainforest}, CSKG provides two additional triples, stating that tropical is an instance of place and that water is related to tropical.\footnote{\texttt{fn:fe:place mw:HasInstance rg:tropical+/c/en/tropical+vg:tropical}} \footnote{\texttt{rg:water+/c/en/water+at:water+vg:water /r/RelatedTo rg:tropical+/c/en/tropical+vg:tropical}}
The first additional edge stems from our mappings from FrameNet to ConceptNet, whereas the second comes from Visual Genome. Interestingly, the above example comes from CSQA, which has been inferred from ConceptNet, showing that most commonsense knowledge is still largely missing in existing resources. We note that, while CSKG increases the coverage with respect to available commonsense knowledge, it is also incomplete: in the above example, useful information such as warm temperatures being typical for tropical rainforests is still missing.


\section{Discussion}
\label{sec:discussion}


The graph metrics in the previous section indicate that CSKG is well-connected, thus showing the impact of our mappings across sources and the merge of identical nodes. Furthermore, the novel evidence brought by CSKG on downstream QA tasks (section \ref{ssec:downstream}) is a signal that can be exploited by reasoning systems to enhance their performance and robustness.
 What are the next steps for CSKG? We discuss three ongoing pursuits.


\textbf{Downstream tasks} Injecting knowledge from ConceptNet has improved the performance of existing question-answering systems like KagNet~\cite{lin2019kagnet}, GapQA \cite{khot2019s}, and HyKAS~\cite{ma2019towards}.\footnote{For an overview of commonsense knowledge sources, reasoners, and benchmarks, see \cite{storks2019commonsense}.} 
At the same time, the error analyses and discussions of these systems reveals that missing knowledge is directly responsible for a portion of their errors. 
We expect that the richer and more diverse knowledge captured by CSKG, as quantified in section \ref{sec:cskg}, would benefit commonsense QA systems. To validate this expectation, we developed a modular evaluation framework, and we computed various graph and word embeddings of CSKG and it subsets. At present, we are running experiments with adaptations of HyKAS and KagNet on six such datasets. 


\textbf{New resources and mappings} We intend to continue integrating the resources listed in section \ref{sec:sources}: WebChild, VerbNet, and CEO. We envision their integration following our approach to be fairly straightforward. Moreover, as machine common sense knowledge is an active research area (section \ref{sec:intro}), we expect additional knowledge sources to be released and integrated in CSKG in the near future. 
Similarly, we intend to create further links between the resources to maximize the connectivity of the ingredients within CSKG.

\textbf{Semantic enrichment} The semantics of CSKG could be improved by refining its relations. For instance, its most common ConceptNet relation, \texttt{/r/RelatedTo} abstracts over various specific predicates, and its full set of relations is unknown and potentially large. In Figure \ref{fig:cskg}, it expresses \textit{containment} (piano \textit{related to} piano key), and \textit{inheritance} (piano key \textit{related to} key), while elsewhere (farmer \textit{related to} man), it obfuscates occupation. Clustering its knowledge into more specific predicates would improve the semantics of CSKG.



\section{Conclusions and Future Work}
\label{sec:conclusions}

The traditional goal of capturing, representing, and leveraging commonsense knowledge has recently gained traction, thanks to initiatives like DARPA's Machine Common Sense~\cite{gunning2018machine}. 
In this paper, we reviewed representative commonsense knowledge sources. While they contain complementary knowledge that would be beneficial as a whole for downstream tasks, such usage is prevented by their different approaches, foci, strenghts, and weaknesses.
Optimizing for simplicity, modularity, and utility, we proposed a property graph that describes many nodes with a few edge types, maximizes the high-quality links across subgraphs, and enables natural language access. We applied this approach to consolidate a commonsense knowledge graph (CSKG) from seven very diverse sources: a text-based commonsense knowledge graph ConceptNet, a general-purpose taxonomy Wikidata, an image description dataset Visual Genome, a procedural knowledge source ATOMIC, and three lexical sources: WordNet, Roget, and FrameNet. It describes 4.7 million nodes with 17.2 million statements. Our analysis showed that CSKG is a well-connected graph and more than `a simple sum of its parts'. On four commonsense QA datasets, it consistently increased the recall of relevant triples by 2-4 times compared to ConceptNet. At present, we are investigating whether this additional signal brought by CSKG helps reasoning on these tasks, and we are integrating further resources reviewed in this paper. CSKG will be released under CC BY-SA 4.0, 
the most permissive license allowed by its components.

\bibliography{sample}
\bibliographystyle{plainnat}

\end{document}